\documentclass[10pt,twocolumn,letterpaper]{article}

\usepackage{iccv}
\usepackage{times}
\usepackage{epsfig}
\usepackage{graphicx}
\usepackage{amsmath}
\usepackage{amssymb}
\usepackage{algorithm}
\usepackage{algorithmic}

\usepackage[pagebackref=true,breaklinks=true,letterpaper=true,colorlinks,bookmarks=false]{hyperref}

\newcommand{\argmin}{\operatornamewithlimits{argmin}}

\iccvfinalcopy 

\ificcvfinal\fi
\begin{document}

\title{Adaptive Low-Rank Kernel Subspace Clustering}
\author{Pan Ji, Ian Reid, Ravi Garg\\
University of Adelaide\\
{\tt\small firstname.lastname@adelaide.edu.au}
\and
Hongdong Li\\
Australian National University\\
{\tt\small hongdong.li@anu.edu.au}
\and Mathieu Salzmann\\
EPFL\\
{\tt\small mathieu.salzmann@epfl.ch}
}

\maketitle

\begin{abstract}
In this paper, we present a kernel subspace clustering method that can handle non-linear models. In contrast to recent kernel subspace clustering methods which use predefined kernels, we propose to adaptively solve a low-rank kernel matrix, with which mapped data in the feature space is not only low-rank but also self-expressive. In this manner, the low-dimensional subspace structures of the (implicitly) mapped data are retained and manifested in the high-dimensional feature space. We evaluate the proposed method extensively on both motion segmentation and image clustering benchmarks, and obtain superior results, outperforming the kernel subspace clustering method that uses standard kernels~\cite{patel2014kernel} and other state-of-the-art linear subspace clustering methods.
\end{abstract}


\section{Introduction}

Subspace clustering denotes the problem of clustering data points drawn from a union of low-dimensional linear (or affine) subspaces into their respective subspaces. This problem has many applications in computer vision, such as motion segmentation and image clustering. To give a concrete example, under an affine camera model, the trajectories of points on a rigidly moving object lie in a linear subspace of dimension up to four; thus motion segmentation 
can be cast as a subspace clustering problem~\cite{elhamifar2009sparse}.

\begin{figure}[!t]
\centering
\includegraphics[width=0.95\linewidth]{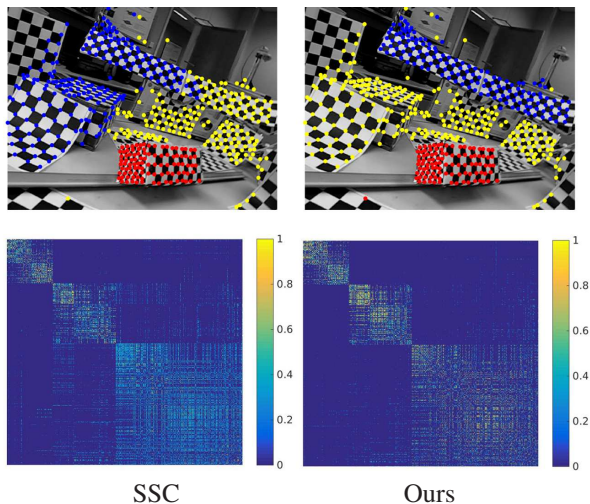}\\
SSC \hspace{2.8cm} Ours
\caption{Top row: motion segmentation results by SSC (left) and our method (right) on the 2RT3RCR sequence of Hopkins155 dataset; feature points from the same motion are marked with the same color. Bottom row: normalized affinity matrices by SSC and our method. While SSC correctly clusters $64.38\%$ of the feature trajectories, our method achieves $98.04\%$ clustering accuracy. Best viewed in color.}
\label{fig1:2RT3RCR}
\end{figure}

Existing subspace clustering methods can be roughly divided into three categories~\cite{vidal2011subspace}: algebraic algorithms, statistical methods, and spectral clustering-based methods. We refer the reader to~\cite{vidal2011subspace} for a comprehensive review of the literature of subspace clustering. Recently, there has been a surge of spectral clustering-based methods~\cite{elhamifar2009sparse,liu2010robust,lu2012robust,wang2013provable,ji2014efficient,ji2015shape,you2016scalable,you2016oracle}, which consist of first constructing an affinity matrix and then applying spectral clustering~\cite{ng2001spectral}. All these methods, however, can only handle linear (or affine) subspaces. In practice, the data points may not fit exactly to a linear subspace model. For example, in motion segmentation~(see Figure~\ref{fig1:2RT3RCR}), the camera often has some degree of perspective distortion so that the affine camera assumption does not hold; in this case, the trajectories of one motion rather lie in a non-linear subspace (or sub-manifold)~\footnote{In this paper, we confine our discussion to data structures that non-linearly deviate from linear subspaces, but are not arbitrarily far from linear subspaces. Therefore, arbitrary manifold clustering~\cite{souvenir2005manifold}, e.g., for Olympic rings and spirals, is out of the scope of this paper.}.

A few other methods~\cite{patel2013latent,patel2014kernel,nguyen2015kernel,xiao2016robust,yin2016kernel} extended linear subspace clustering to nonlinear counterparts by exploiting the kernel trick. In particular,~\cite{patel2013latent,patel2014kernel} kernelized sparse subspace clustering (SSC)~\cite{elhamifar2009sparse,elhamifar2013sparse} by replacing the inner product of the data matrix with the polynomial kernel or Gaussian RBF kernel matrices; similarly,~\cite{nguyen2015kernel,xiao2016robust} kernelized the method of low rank representation (LRR)~\cite{liu2010robust,liu2013robust}. \cite{yin2016kernel} assumed that data points were drawn from symmetric positive definite (SPD) matrices and applied the Log-Euclidean kernel on SPD matrices to kernelize SSC. However, with the {\it pre-defined} kernels used in all these methods, the data after (implicit) mapping to the feature space have no guarantee to be low-rank, and thus are very unlikely to form multiple low-dimensional subspace structures in the feature space.

In this paper, by contrast, we propose a joint formulation to adaptively solve (in an unsupervised manner) a low-rank kernel mapping 
such that the data in the resulting feature space is both low-rank and self-expressive. Intuitively, enforcing the kernel feature mapping to be low-rank will encourage the data to form linear subspace structures in feature space. Our idea of low-rank kernel mapping is general and could, in principle, be implemented within most self-expressiveness-based subspace clustering frameworks~\cite{elhamifar2009sparse,liu2010robust,lu2012robust,wang2013provable}. Here, in particular, we make use of the SSC one of~\cite{elhamifar2009sparse}. This allows us to make use of the Alternating Direction Method of Multipliers (ADMM)~\cite{lin2010augmented,boyd2011distributed} to derive an efficient solution to the resulting optimization problem.

We extensively evaluate our method on multiple motion segmentation and image clustering datasets, and show that it significantly outperforms the linear subspace clustering methods of~\cite{elhamifar2013sparse,vidal2014low,liu2013robust,you2016scalable} and the method of~\cite{patel2014kernel} based on pre-defined kernels. Specifically, we achieve state-of-the-art results on the Hopkins155 motion segmentation dataset~\cite{tron2007benchmark}, the Extended Yale B face image clustering dataset~\cite{lee2005acquiring}, the ORL face image clustering dataset~\cite{samaria1994parameterisation}, and the COIL-100 image clustering dataset~\cite{nenecolumbia}.

\section{Subspace Self-Expressiveness}

Modern subspace clustering methods rely on building an affinity matrix such that data points from the same subspace have high affinity values and those from different subspaces have low affinity values (ideally zero). Recent self-expressiveness-based methods~\cite{elhamifar2013sparse,liu2013robust,you2016oracle} resort to the so-called subspace self-expressiveness property, \ie, one point from one subspace can be represented as a linear combination of other points in the same subspace, and leverage the self-expression coefficient matrix as the affinity matrix for spectral clustering.

Specifically, given a data matrix ${\bf X}\in\mathbb{R}^{D\times N}$ (with each column a data point), subspace self-expressiveness means that one can express ${\bf X} = {\bf X}{\bf C}$, where ${\bf C}$ is the self-expression coefficient matrix. As shown in~\cite{ji2014efficient}, under the assumption of subspaces being independent, the optimal solution for ${\bf C}$ obtained by minimizing certain norms of ${\bf C}$ has a block-diagonal structure (up to permutations), \ie, $c_{ij}\neq 0$ only if points $i$ and $j$ are from the same subspace. In other words, we can address the subspace clustering problem by solving the following optimization problem
\begin{equation}
\min\limits_{\bf C} \|{\bf C}\|_p \quad{\rm s.t.}\quad {\bf X} = {\bf X}{\bf C},\; ({\rm diag}({\bf C}) = {\bf 0})\;,
\end{equation}
where $\|\cdot\|_p$ denotes an arbitrary matrix norm, and the constraint ${\rm diag}({\bf C}) = {\bf 0}$~\footnote{${\rm diag}({\rm C})$ denotes the diagonal matrix whose diagonal elements are the same as those on the diagonal of ${\bf C}$.} prevents the trivial identity solution for sparse norms of ${\bf C}$~\cite{elhamifar2013sparse}. In the literature, various norms on ${\bf C}$ have been used to regularize subspace clustering, such as the $\ell_1$ norm in~\cite{elhamifar2009sparse,elhamifar2013sparse}, the nuclear norm in~\cite{liu2010robust,liu2013robust,favaro2011closed,vidal2014low}, the $\ell_2$ norm in~\cite{lu2012robust,ji2014efficient}, a structured norm in~\cite{li2015structured}, and a mixture of $\ell_1$ and $\ell_2$ norms in~\cite{wang2013provable,you2016oracle}.

Compared to another line of research based on local higher-order models~\cite{chen2009spectral,jain2013efficient,purkait2016clustering}, where affinities are constructed from the residuals of {\it local} subspace model fitting, the self-expressiveness-based methods build {\it holistic} connections (or affinities) for all points, in a single, principled optimization problem.
Moreover, this formulation is convex (after certain relaxations), which guarantees globally-optimal solutions. Unfortunately, subspace self-expressiveness only holds for linear (or affine) subspaces. In the following section, we show how to jointly solve a low-rank kernel for non-linear subspace clustering within the framework of self-expressiveness-based subspace clustering, and derive efficient solutions for the resulting formulations.

\section{Low-Rank Kernel Subspace Clustering}

\begin{figure}[!t]
\centering
\includegraphics[width=0.9\linewidth]{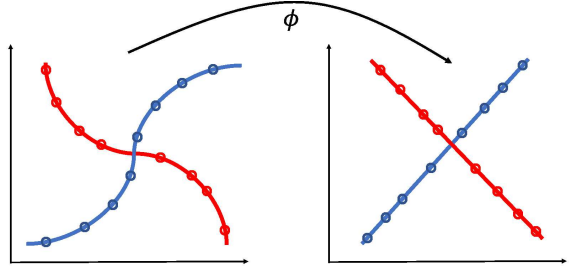}
\caption{Subspace-aware feature mapping: non-linear subspaces (or sub-manifolds) are mapped to linear ones in high-dimensional feature space. Note that we plot this in 2D for the visualization convenience. In fact, the ambient dimension is very high in feature space, and the intrinsic dimension can also be higher than in the input space.}
\label{fig2:kernel_mapping}
\end{figure}

Kernel methods map data points to a high-dimensional (or even infinite dimensional) feature space where linear pattern analysis can be done, corresponding to non-linear analysis in the input data space~\cite{shawe2004kernel}. Instead of explicitly computing the coordinates in the high-dimensional feature space, common practice in kernel methods consists of using the ``kernel trick'', where feature mapping is implicit and inner products between pair of data points in the feature space are computed as kernel values. While the ``kernel trick'' is relatively computationally cheap, for commonly used kernels such as the Gaussian RBF,
we don't know explicitly how the data points are mapped to feature space. Specifically, in the context of subspace clustering, it is very likely that, after an implicit feature mapping, we don't have the desired low-dimensional linear subspace structure in the feature space.

\subsection{Problem Formulation}

In this work, we aim to solve a low-rank kernel mapping that projects the data into a high-dimensional Hilbert space where it has the structure of linear subspaces (see Figure~\ref{fig2:kernel_mapping}). While our approach is general, we implement it within the sparse subspace clustering (SSC) framework~\cite{elhamifar2009sparse,elhamifar2013sparse}, formulated as
\begin{equation}
\label{eq:ssc}
\min\limits_{\bf C} \|{\bf C}\|_1 \quad{\rm s.t.}\quad {\bf X} = {\bf X}{\bf C},\; {\rm diag}({\bf C}) = {\bf 0}\;,
\end{equation} 
which seeks a sparse representation of every data point using the other points as the dictionary.

When the structure of linear subspaces is present in the Hilbert feature space, the feature mapping $\phi({\bf X})$ should have low rank. Since we would like the data in  Hilbert space to still lie on multiple linear subspaces, we also expect it to be self-expressive. Combining these properties leads to the optimization problem as follows
\begin{equation}
\begin{split}
&\min\limits_{{\bf K},{\bf C}}\;\; \|\phi({\bf X})\|_* + \lambda\|{\bf C}\|_1 \\
{\rm s.t.}\;\; 
\label{eq:lrksc1}
&\phi({\bf X}) = \phi({\bf X}){\bf C}\;,\; {\rm diag}({\bf C}) = {\bf 0}\;, 
\end{split}
\end{equation}
where $\phi(\cdot)$ is an unknown kernel mapping function and $\lambda$ is a tradeoff parameter. Here, we minimize the nuclear norm $\|\phi({\bf X})\|_*$, which is a convex surrogate of ${\rm rank}(\phi({\bf X}))$, to encourage $\phi({\bf X})$ to have low rank.  
In practice, the data points often contain noise. Therefore, we can relax the equality constraint in~\eqref{eq:lrksc1} and make it a regularization term in the objective, \ie, $\|\phi({\bf X}) - \phi({\bf X}){\bf C}\|_F^2$. In kernel methods, we normally don't know the explicit form of $\phi(\cdot)$, so we need to apply the ``kernel trick''. To this end, we can then expand this regularization term and have 
\begin{equation}
\label{eq:kernel}
\|\phi({\bf X}) - \phi({\bf X}){\bf C}\|_F^2 = {\rm tr}({\bf K}-2{\bf K}{\bf C}+{\bf C}^T{\bf K}{\bf C})\;,
\end{equation}
which does not explicitly depend on $\phi({\bf X})$ anymore, but on the kernel Gram matrix ${\bf K} = \phi({\bf X})^T\phi({\bf X})$.

However, there are still two hurdles in optimizing~\eqref{eq:lrksc1}: ({\bf i}) the first term ($\|\phi({\bf X})\|_*$) in the objective depends on the kernel mapping $\phi({\bf X})$ which does not have explicit form in most kernel methods; ({\bf ii}) the kernel mapping function in this formulation is under-constrained in the sense that one can always map the data to all zero to achieve the minimum value of the objective. 

For the first hurdle, one may think of minimizing $\|{\bf K}\|_*$ instead of $\|\phi({\bf X})\|_*$ as the rank of ${\bf K}$ is equal to the rank of $\phi({\bf X})$. However, minimizing $\|{\bf K}\|_*$ will not lead to a low-rank $\phi({\bf X})$, because we have $\|{\bf K}\|_* = {\rm tr}\sqrt{{\bf K}^T{\bf K}} = {\rm tr}\big(\phi({\bf X})^T\phi({\bf X})\big) = \|\phi({\bf X})\|_F^2$. So minimizing $\|{\bf K}\|_*$ is equivalent to minimizing $\|\phi({\bf X})\|_F^2$, which doesn't encourage the data in feature space to have low rank~\cite{garg2016non}. Fortunately, it has been shown in~\cite{garg2016non} that this hurdle can be circumvented by using a re-parametrization, which leads to a closed form solution for robust rank minimization in the feature space. Since the kernel matrix ${\bf K}$ is symmetric positive semi-definite, we can factorize it as ${\bf K} = {\bf B}^T{\bf B}$, where ${\bf B}$ is a square matrix. It is easy to show that 
\begin{equation}
\|{\bf B}\|_* = \|\phi({\bf X})\|_*\;, \forall\; {\bf B}:\; {\bf K} = {\bf B}^T{\bf B}\;.
\end{equation}
Thus, we can replace $\|\phi({\bf X})\|_*$ with $\|{\bf B}\|_*$ in the objective of \eqref{eq:lrksc1}. 

For the second hurdle, to regularize the kernel mapping, we further enforce our adaptive kernel matrix ${\bf B}^T{\bf B}$ to be close to a pre-defined kernel matrix ${\bf K}_G$ using user-specified kernels.
With an additional affine constraint for affine subspaces, our optimization problem translates to
\begin{equation}
\begin{split}
\min\limits_{{\bf B},{\bf C}}\;\; &\|{\bf B}\|_* + \lambda_1\|{\bf C}\|_1 +\frac{\lambda_2}{2}\|\phi({\bf X}) - \phi({\bf X}){\bf C}\|_F^2 \\
\label{eq:lrksc2}
&\qquad\;+ \frac{\lambda_3}{2}\|{\bf K}_G - {\bf B}^T{\bf B}\|_F^2\;,\\
&{\rm s.t.}\;\; {\bf 1}^T{\bf C} = {\bf 1}^T\;,\;{\rm diag}({\bf C}) = {\bf 0}\;,
\end{split}
\end{equation}
where $\phi({\bf X})^T\phi({\bf X}) = {\bf B}^T{\bf B}$ and ${\bf 1}$ is an all-one column vector. 
The idea of our formulation is that we want to solve an adaptive kernel matrix ${\bf K} = {\bf B}^T{\bf B}$ such that: 
\begin{itemize}\setlength\itemsep{-0.1em}
\item the mapped data in the feature space has low rank; 
\item the unknown kernel matrix (to be solved) is not arbitrary but close to a predefined kernel matrix (\eg, polynomial kernels); 
\item the data points in the feature space still form a multiple linear subspace structure, and are thus self-expressive.
\end{itemize}

To handle the diagonal constraint on ${\bf C}$, we introduce an auxiliary variable ${\bf A}$ such that ${\bf A} = {\bf C} - {\rm diag}({\bf C})$ as in~\cite{elhamifar2013sparse}. Substituting Eq.~\eqref{eq:kernel} into~\eqref{eq:lrksc2}, we have the following equivalent formulation
\begin{equation}
\begin{split}
\min\limits_{{\bf B},{\bf C},{\bf A}}\;\; &\|{\bf B}\|_* + \lambda_1\|{\bf C}\|_1 +\frac{\lambda_2}{2}{\rm tr}({\bf I}-2{\bf A}+\\
\label{eq:lrksc3}
&{\bf A}{\bf A}^T){\bf B}^T{\bf B}+ \frac{\lambda_3}{2}\|{\bf K}_G - {\bf B}^T{\bf B}\|_F^2\;,\\
{\rm s.t.}\;\; &{\bf A} = {\bf C} - {\rm diag}({\bf C})\;,\;{\bf 1}^T{\bf A} = {\bf 1}^T\;.
\end{split}
\end{equation}

Below, we show how to solve this problem efficiently.

\subsection{Solutions via ADMM}
The above optimization problem is non-convex (or bi-convex) due to the bilinear terms in the objective. Here we propose to solve it via the Alternating Direction Method of Multipliers (ADMM)~\cite{lin2010augmented,boyd2011distributed}.
Recently, the ADMM has gained popularity to solve non-convex problems~\cite{li2015global}, especially bilinear ones~\cite{del2010bilinear,ji2014robust,shen2014augmented}. A convergence analysis of the ADMM for certain non-convex problems is provided in~\cite{hong2016convergence}. We will also give our empirical convergence analysis in the next section.

To derive the ADMM solution to~\eqref{eq:lrksc3}, we first need to compute its augmented Lagrangian. This is given by
\begin{equation}
\label{eq:lagragian1}
\begin{split}
\mathcal{L}({\bf B},{\bf C},{\bf A},{\bf Y}_1,{\bf y}_2) = \|{\bf B}\|_* + \lambda_1\|{\bf C}\|_1 + 
\frac{\lambda_2}{2}{\rm tr}\\({\bf I}-2{\bf A}+{\bf A}{\bf A}^T){\bf B}^T{\bf B} + \frac{\lambda_3}{2}\|{\bf K}_G - {\bf B}^T{\bf B}\|_F^2 + \\
{\rm tr}\; {\bf Y}^T_1({\bf A}- {\bf C} + {\rm diag}({\bf C})) + {\rm tr}\; {\bf y}^T_2({\bf 1}^T{\bf A} - {\bf 1}^T) + \\
\frac{\rho}{2}\big(\|{\bf A}- {\bf C} + {\rm diag}({\bf C})\|_F^2 + \|{\bf 1}^T{\bf A} - {\bf 1}^T\|_2^2\big)\;,
\end{split}
\end{equation}
where ${\bf Y}_1\in\mathbb{R}^{N\times N}$ and ${\bf y}_2\in\mathbb{R}^{1\times N}$ are the Lagrange multipliers corresponding to the equality constraints in~\eqref{eq:lrksc3}, and $\rho$ is the penalty parameter for the augmentation term in the Lagrangian.

The ADMM works in an iterative manner by updating one variable while fixing the other ones~\cite{lin2010augmented} , \ie,
\begin{subequations}
\begin{align}
{\bf C}^* &= \argmin_{\bf C} \mathcal{L}({\bf B},{\bf C},{\bf A},{\bf Y}_1,{\bf y}_2)\;,\\
{\bf A}^* &= \argmin_{\bf A} \mathcal{L}({\bf B},{\bf C},{\bf A},{\bf Y}_1,{\bf y}_2)\;,\\
{\bf B}^* &= \argmin_{\bf B} \mathcal{L}({\bf B},{\bf C},{\bf A},{\bf Y}_1,{\bf y}_2)\;,
\end{align}
\end{subequations}
and then updating the Lagrange multipliers ${\bf Y}_1, {\bf y}_2$.

\noindent {\bf (1) Updating ${\bf C}$}

The update of ${\bf C}$ can be achieved by solving the following subproblem
\begin{equation}
\label{updateC}
\min\limits_{\bf C} \frac{\lambda_1}{\rho}\|{\bf C}\|_1 + \frac{1}{2}\|{\bf C} - {\rm diag}({\bf C})-({\bf A} + {\bf Y}_1/\rho)\|_F^2\;.
\end{equation}
This subproblem has a closed-from solution given by
\begin{equation}
\label{eq:updateC-solution}
{\bf C}^* = {\bf J} - {\rm diag}({\bf J})\;,
\end{equation}
where ${\bf J} = \mathcal{T}_{\frac{\lambda_1}{\rho}}({\bf A} + {\bf Y}_1/\rho)$, and $\mathcal{T}$ is an element-wise soft-thresholding operator defined as $\mathcal{T}_{\tau}(x) = {\rm sign}(x)\cdot\max(|x|-\tau,0)$.

\noindent {\bf (2) Updating ${\bf A}$}

To update ${\bf A}$, we must solve the subproblem 
\begin{align}
\label{updateA}
\min\limits_{\bf A} &\frac{\lambda_2}{2}{\rm tr}(-2{\bf A}+{\bf A}{\bf A}^T){\bf B}^T{\bf B} + {\rm tr}\;{\bf Y}^T_1{\bf A} + {\rm tr}\; {\bf y}^T_2{\bf 1}^T{\bf A}\notag\\
+&\frac{\rho}{2}\big(\|{\bf A}- {\bf C} + {\rm diag}({\bf C})\|_F^2 + \|{\bf 1}^T{\bf A} - {\bf 1}^T\|_2^2\big)\;.
\end{align}
This can be achieved by taking the derivative w.r.t. ${\bf A}$, and setting it to zero. This again yields a closed-form solution given by
\begin{equation}
\begin{split}
\label{eq:updateA-solution}
{\bf A}^* = \big(\lambda_2{\bf B}^T{\bf B} + \rho({\bf I} + {\bf 1}{\bf 1}^T)\big)^{-1}\big( \lambda_2{\bf B}^T{\bf B} - {\bf Y}_1 \\
- {\bf 1}{\bf y}_2 + \rho({\bf C} - {\rm diag}({\bf C}) + {\bf 1}{\bf 1}^T)\big)\;,
\end{split}
\end{equation}
where ${\bf I}$ is an identity matrix.

\noindent {\bf (3) Updating ${\bf B}$}

${\bf B}$ can be updated by solving the following subproblem
\begin{equation}
\label{updateB}
\min\limits_{\bf B} \|{\bf B}\|_* + \frac{\lambda_3}{2}\big\|{\bf B}^T{\bf B} - \tilde{\bf K}_G\big\|_F^2\;,
\end{equation}
with $\tilde{\bf K}_G = {\bf K}_G - \frac{\lambda_2}{2\lambda_3}({\bf I} - 2{\bf A}^T + {\bf A}{\bf A}^T)$.~\footnote{In our implementation, we make $\tilde{\bf K}_G$ a symmetric matrix by computing $\frac{1}{2}(\tilde{\bf K}_G+\tilde{\bf K}_G^T)$.}
Fortunately, this subproblem also has a closed-form solution given by
\begin{equation}
\label{eq:updateB-solution}
{\bf B}^* = \Gamma^*{\bf V}^T\;,
\end{equation}
where $\Gamma^*$ and ${\bf V}$ are both related to the singular value decomposition (SVD) of $\tilde{\bf K}_G$. Let $\tilde{\bf K}_G = {\bf U}\Sigma{\bf V}^T$ denote the SVD of $\tilde{\bf K}_G$. $\Gamma^*$ is a diagonal matrix, \ie, $\Gamma^* = {\rm diag}(\gamma_1^*,\cdots,\gamma_N^*)$, with $\gamma_i^* = \argmin_{\gamma_i\geq0} \frac{\lambda_3}{2}(\sigma_i-\gamma_i^2)^2 + \gamma_i$ (\ie, $\gamma_i^*\in\{x\in\mathbb{R}_+ | x^3 - \sigma_ix + \frac{1}{2\lambda_3} = 0\}\cup\{0\}$), where $\sigma_i$ is the $i^{\rm th}$ singular value of $\tilde{\bf K}_G$ (see~\cite{garg2016non} for a complete proof of this result). In other words, $\Gamma^*$ can be obtained by first solving a set of depressed cubic equations whose first-order coefficients come from the singular values of $\tilde{\bf K}_G$ and then select the non-negative solution that minimizes $f(\gamma_i) = \frac{\lambda_3}{2}(\sigma_i-\gamma_i^2)^2 + \gamma_i$; ${\bf V}$ is the matrix containing the singular vectors of $\tilde{\bf K}_G$. Note that the solution to~\eqref{updateB} is non-unique as one can multiply an arbitrary orthogonal matrix to the left of~\eqref{eq:updateB-solution} without changing the objective value in~\eqref{updateB}. However, this non-uniqueness does not affect the final clustering as the solution for C remains the same.

The algorithm for solving~\eqref{eq:lrksc3} via the ADMM is outlined in Algorithm~\ref{Algorithm1}.

\begin{algorithm}[t!]
\caption{Solving~\eqref{eq:lrksc3} via the ADMM}
\label{Algorithm1}
\begin{algorithmic}
\REQUIRE 
Data matrix ${\bf X}$, weight parameters $\lambda_1$, $\lambda_2$, $\lambda_3$
 \\ \vspace{0.1cm}
\hspace{-0.35cm}{\bf Initialize:} ${\bf B} = \sqrt{{\bf K}_G}$ (where ${\bf K}_G$ is computed with a user-specified kernel), ${\bf C} = {\bf 0}$, ${\bf A} = {\bf 0}$ , ${\bf Y}_1 = {\bf 0}$, ${\bf y}_2 = {\bf 0}$, $\rho = 10^{-8}$, $\rho_m = 10^{10}$, $\eta = 20$, $\epsilon = 10^{-6}$\\ \vspace{0.2cm}
\WHILE {not converged}
\STATE 1.~Sequentially update ${\bf C}$, ${\bf A}$, ${\bf B}$ in closed-from by \eqref{eq:updateC-solution}, \eqref{eq:updateA-solution}, \eqref{eq:updateB-solution} respectively;

\STATE 2.~Update Lagrange multipliers and penalty variables as follows
{\small
\begin{align*}
{\bf Y}_1 &:= {\bf Y}_1 + \rho({\bf A}- {\bf C} + {\rm diag}({\bf C}))\;,\\
{\bf y}_2 &:= {\bf y}_2 + \rho({\bf 1}^T{\bf A} - {\bf 1}^T)\;,\\
\rho      &:= \min(\eta\rho,\rho_m)\;;
\end{align*}
}

\vspace{-0.5cm}
\STATE 3. Check the convergence conditions \\$\|{\bf A}- {\bf C} + {\rm diag}({\bf C})\|_{\infty} \leq \epsilon$, and $\|{\bf 1}^T{\bf A} - {\bf 1}^T\|_{\infty} \leq \epsilon$; \\
\ENDWHILE
\vspace{0.2cm}
\ENSURE ~Coefficient matrix ${\bf C}$
\end{algorithmic}
\end{algorithm}

\subsection{Handling Gross Corruptions}
Our formulation in~\eqref{eq:lrksc2} can be sensitive to gross data corruptions (\eg, Laplacian noise) due to the $\ell_2$ norm regularization. When data points are grossly contaminated, we need to model the gross errors in the kernel matrix. To this end, we assume that the gross corruptions in the data are sparse so that we can model them with an $\ell_1$ regularizer~\footnote{In principle, one can also model the gross corruptions with structured sparse norms such as $\ell_{2,1}$ norm if some data points are completely outliers. Here we use the $\ell_1$ norm regularizer because we assume that gross corruptions only happen to some sparse entries of the data vectors.}. This lets us derive the following formulation
\begin{equation}
\label{eq:lrksc4}
\begin{split}
\min\limits_{{\bf B},{\bf C},{\bf A}, {\bf E}}\;\; &\|{\bf B}\|_* + \lambda_1\|{\bf C}\|_1 + \frac{\lambda_2}{2}\|\phi({\bf X}) - \\
&\phi({\bf X}){\bf A}\|_F^2 + \lambda_3\|{\bf E}\|_1\\
{\rm s.t.}\;\; &{\bf A} = {\bf C} - {\rm diag}({\bf C}),\;{\bf 1}^T{\bf A} = {\bf 1}^T\;,\\
&{\bf K}_G = {\bf B}^T{\bf B} + {\bf E}\;,
\end{split}
\end{equation} 
where $\phi({\bf X})^T\phi({\bf X}) = {\bf B}^T{\bf B}$, and where we decompose the predefined kernel matrix ${\bf K}_G$ into the sum of a low-rank kernel matrix ${\bf B}^T{\bf B}$ and a sparse outlier term ${\bf E}$.

Similarly to~\eqref{eq:lrksc3}, we can again solve~\eqref{eq:lrksc4} with the ADMM. To this end, we derive its augmented Lagrangian as
{\small
\begin{align}
&\mathcal{L}({\bf B},{\bf C},{\bf A},{\bf E},{\bf Y}_1,{\bf y}_2,{\bf Y}_3) = \|{\bf B}\|_* + \lambda_1\|{\bf C}\|_1\notag \\ 
& +\frac{\lambda_2}{2}{\rm tr}({\bf I}-2{\bf A}+{\bf A}{\bf A}^T){\bf B}^T{\bf B} + \lambda_3\|{\bf E}\|_1  \notag\\
&+{\rm tr}{\bf Y}^T_1({\bf A}- {\bf C} + {\rm diag}({\bf C})) + {\rm tr}\; {\bf y}^T_2({\bf 1}^T{\bf A} - {\bf 1}^T) \notag\\ 
&+{\rm tr}{\bf Y}_3^T({\bf K}_G - {\bf B}^T{\bf B}-{\bf E}) + 
\frac{\rho}{2}\big(\|{\bf A}- {\bf C} + {\rm diag}({\bf C})\|_F^2 \notag\\ 
&+ \|{\bf 1}^T{\bf A} - {\bf 1}^T\|_2^2 + \|{\bf K}_G - {\bf B}^T{\bf B}-{\bf E}\|_F^2\big)\;.
\end{align}}
We can then derive the subproblems to update each of the variables by minimizing $\mathcal{L}({\bf B},{\bf C},{\bf A},{\bf E},{\bf Y}_1,{\bf y}_2,{\bf Y}_3)$.

\begin{algorithm}[t!]
\caption{Solving~\eqref{eq:lrksc4} via the ADMM}
\label{Algorithm2}
\begin{algorithmic}
\REQUIRE 
Data matrix ${\bf X}$, weight parameters $\lambda_1$, $\lambda_2$, $\lambda_3$
 \\ \vspace{0.1cm}
\hspace{-0.35cm}{\bf Initialize:} ${\bf B} = \sqrt{{\bf K}_G}$ (where ${\bf K}_G$ is computed with a user-specified kernel), ${\bf C} = {\bf 0}$, ${\bf A} = {\bf 0}$, ${\bf E} = {\bf 0}$, ${\bf Y}_1 = {\bf 0}$, ${\bf y}_2 = {\bf 0}$, ${\bf Y}_3 = {\bf 0}$, $\rho = 10^{-8}$, $\rho_m = 10^{10}$, $\eta = 20$, $\epsilon = 10^{-6}$\\ \vspace{0.2cm}
\WHILE {not converged}
\STATE 1.~Sequentially update ${\bf C}$, ${\bf A}$, ${\bf B}$, ${\bf E}$  in closed-form  by ~\eqref{eq:updateC-solution}, \eqref{eq:updateA-solution}, \eqref{eq:updateB-solution}, \eqref{eq:updateE-solution} respectively;

\STATE 2.~Update Lagrange multipliers and penalty variables as follows
{\small
\begin{align*}
{\bf Y}_1 &:= {\bf Y}_1 + \rho({\bf A}- {\bf C} + {\rm diag}({\bf C}))\;,\\
{\bf y}_2 &:= {\bf y}_2 + \rho({\bf 1}^T{\bf A} - {\bf 1}^T)\;,\\
{\bf Y}_3 &:= {\bf Y}_3 + \rho({\bf K}_G - {\bf B}^T{\bf B} - {\bf E})\\
\rho      &:= \min(\eta\rho,\rho_m)\;;
\end{align*}
}

\vspace{-0.5cm}
\STATE 3. Check the convergence conditions \\$\|{\bf A}- {\bf C} + {\rm diag}({\bf C})\|_{\infty} \leq \epsilon$, $\|{\bf 1}^T{\bf A} - {\bf 1}^T\|_{\infty} \leq \epsilon$, and $\|{\bf K}_G - {\bf B}^T{\bf B} - {\bf E}\|_{\infty} \leq \epsilon$; \\
\ENDWHILE
\vspace{0.2cm}
\ENSURE ~Coefficient matrix ${\bf C}$
\end{algorithmic}
\end{algorithm}

\noindent {\bf (1) Updating ${\bf C}$, ${\bf A}$} and ${\bf B}$

The subproblems for updating ${\bf C}$ and ${\bf A}$ are exactly the same as in~\eqref{updateC} and~\eqref{updateA}. Correspondingly, the solutions are also the same as in~\eqref{eq:updateC-solution} and~\eqref{eq:updateA-solution}. The subproblem to update ${\bf B}$ is similar to~\eqref{updateB}, except that $\tilde{\bf K}_G$ is now defined as
\begin{equation}
\tilde{\bf K}_G =  {\bf K}_G - \frac{1}{\rho}(\frac{\lambda_2}{2}({\bf I}-2{\bf A}^T+{\bf A}{\bf A}^T)-{\bf Y}_3) - {\bf E}\;.
\end{equation}

\noindent {\bf (2) Updating ${\bf E}$}

The subproblem to update ${\bf E}$ can be written as 
\begin{equation}
\label{updateE}
\min_{\bf E} \frac{\lambda_3}{\rho}\|{\bf E}\|_1 + \frac{1}{2}\|{\bf E} - ({\bf K}_G-{\bf B}^T{\bf B}+{\bf Y}_3/\rho )\|_F^2\;,
\end{equation}
which has a closed-form solution given by
\begin{equation}
\label{eq:updateE-solution}
{\bf E}^* = \mathcal{T}_{\frac{\lambda_3}{\rho}}({\bf K}_G-{\bf B}^T{\bf B}+{\bf Y}_3/\rho)\;.
\end{equation}

The algorithm for solving~\eqref{eq:lrksc4} with the ADMM is outlined in Algorithm~\ref{Algorithm2}.

\subsection{The Complete Algorithm}

Given the data matrix ${\bf X}$, we solve either~\eqref{eq:lrksc3} or~\eqref{eq:lrksc4} with Algorithm~\ref{Algorithm1} or~\ref{Algorithm2}, respectively, depending on whether the data points are grossly contaminated or not. After we get the coefficient matrix ${\bf C}$, we then construct the affinity matrix with an extra normalization step on ${\bf C}$ as in SSC~\cite{elhamifar2013sparse}. Finally, we apply the spectral clustering algorithm~\cite{ng2001spectral,shi2000normalized} to get the clustering results.
Our complete algorithm for low-rank kernel subspace clustering (LRKSC) is outlined in Algorithm~\ref{Algorithm3}.

\section{Experiments}

We compare our method with the following baselines: low rank representation (LRR)~\cite{liu2013robust}, sparse subspace clustering (SSC)~\cite{elhamifar2013sparse}, kernel sparse subspace clustering (KSSC)~\cite{patel2014kernel}, low rank subspace clustering (LRSC)~\cite{vidal2014low}, and sparse subspace clustering by orthogonal matching pursuit (SSC-OMP)~\cite{you2016scalable}. Specifically, KSSC is the kernelized version of SSC, and SSC-OMP is a scalable version of SSC.~\footnote{While there are other kernel subspace clustering methods~\cite{xiao2016robust,yin2016kernel} in the literature, their source codes are not yet publicly available. So we are unable to compare with their results. However, our comparison with KSSC already shows the benefits of solving for a low-rank kernel over a pre-defined kernel.} The metric for quantitative evaluation is the ratio of wrongly clustered points, \ie,
\begin{equation*}
{\rm Err}\;\% = \frac{\#\; \text{of wrongly clustered points}}{\text{total $\#$ of points}}\times 100\%\;.
\end{equation*}
For LRR, SSC, LRSC and SSC-OMP, we used the source codes released by the authors. For KSSC, we used the same ADMM framework as ours (and set $\eta = 3$ for best performance), and update the variables according to the descriptions in the original paper~\cite{patel2014kernel}.

\subsection{Kernel Selection}

For all kernel methods, kernel selection still remains an open but important problem since there is a vast range of possible kernels in practice. The selection of kernels depends on our tasks at hand, the prior knowledge about the data and the types of patterns we expect to discover~\cite{shawe2004kernel}. In the supervised setting, one can possibly use a validation set to choose the kernels that give the best (classification) results. However, in the unsupervised setting, it is hard to define a measure of "goodness" to guide the kernel selection. Nonetheless, in the case of subspace clustering, we can make use of our prior knowledge (or assumption) 
that the data points are not too far away from linear subspaces, and rely on simple kernels to define ${\bf K}_G$. In this paper, we advocate the use of polynomial kernels for kernel subspace clustering, and argue that more complex kernels such as the Gaussian RBF kernels would destroy the subspace structure in the feature space to the extent that even with our proposed adaptive kernel approximation the subspace structure cannot be restored. Specifically, in all our experiments, we use the polynomial kernels $\kappa({\bf x}_1,{\bf x}_2) = ({\bf x}_1^T{\bf x}_2+a)^b$. The degree parameter $b$ is set to either 2 or 3, and the bias parameter $a$ is selected from $[0,15]$.

\subsection{Convergence Analysis}

\begin{algorithm}[!t]
\caption{Adaptive low-rank kernel subspace clustering}
\label{Algorithm3}
\begin{algorithmic}
\REQUIRE 
Data matrix ${\bf X}$, number of subspaces $K$ \\ 
\vspace{0.1cm}

\STATE 1.~Obtain the self-expression coefficient matrix ${\bf C}$ via Algorithm~\ref{Algorithm1} or Algorithm~\ref{Algorithm2};

\STATE 2.~Construct an affinity matrix from ${\bf C}$, and apply spectral clustering with the affinity matrix to get the subspace clustering labels.

\vspace{0.1cm}
\ENSURE ~Subspace clustering labels
\end{algorithmic}
\end{algorithm}

\begin{figure}[!t]
\centering
\begin{tabular}{cc}
\hspace{-0.2cm}\includegraphics[width=0.48\linewidth]{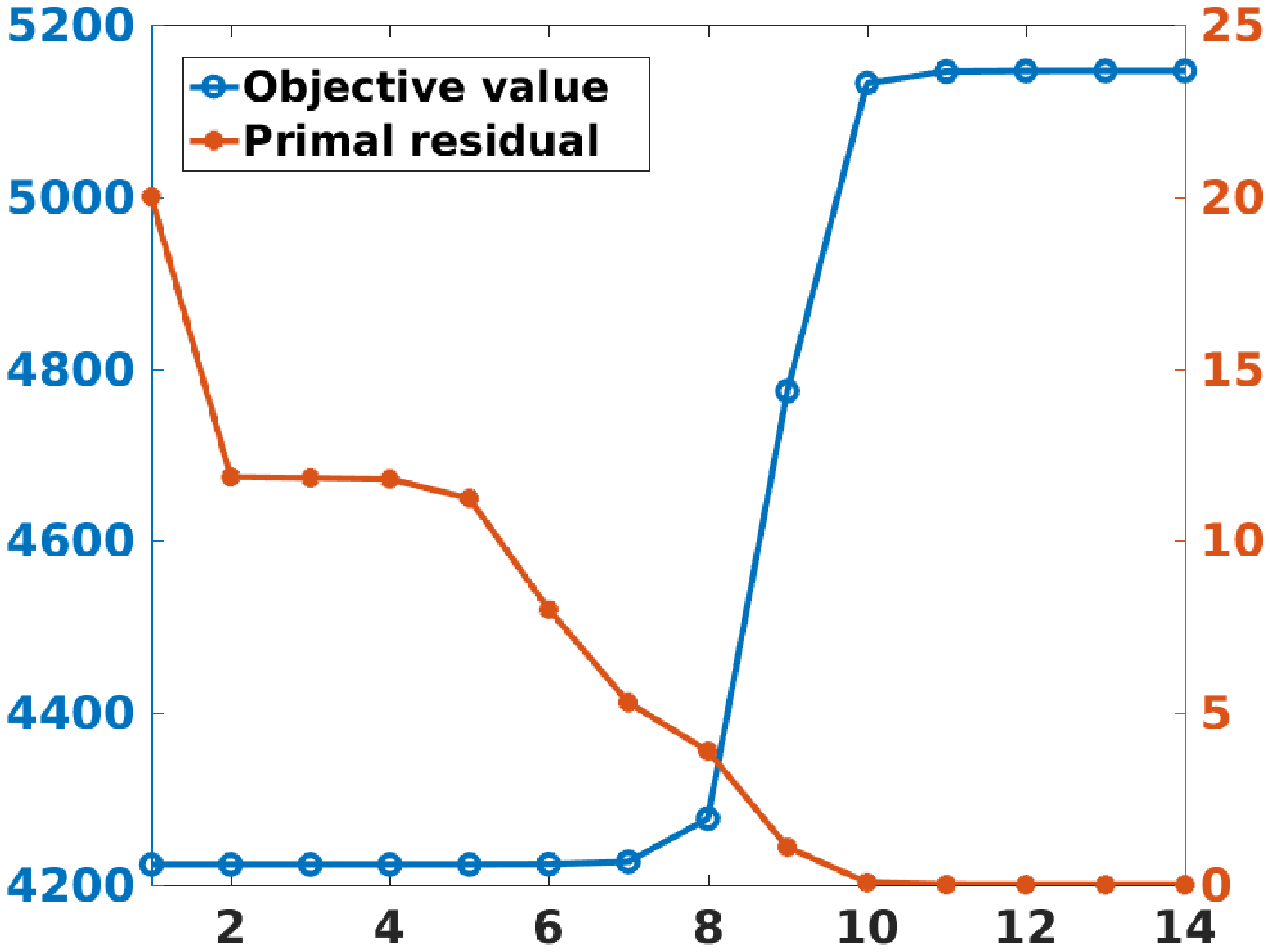}&
\includegraphics[width=0.48\linewidth]{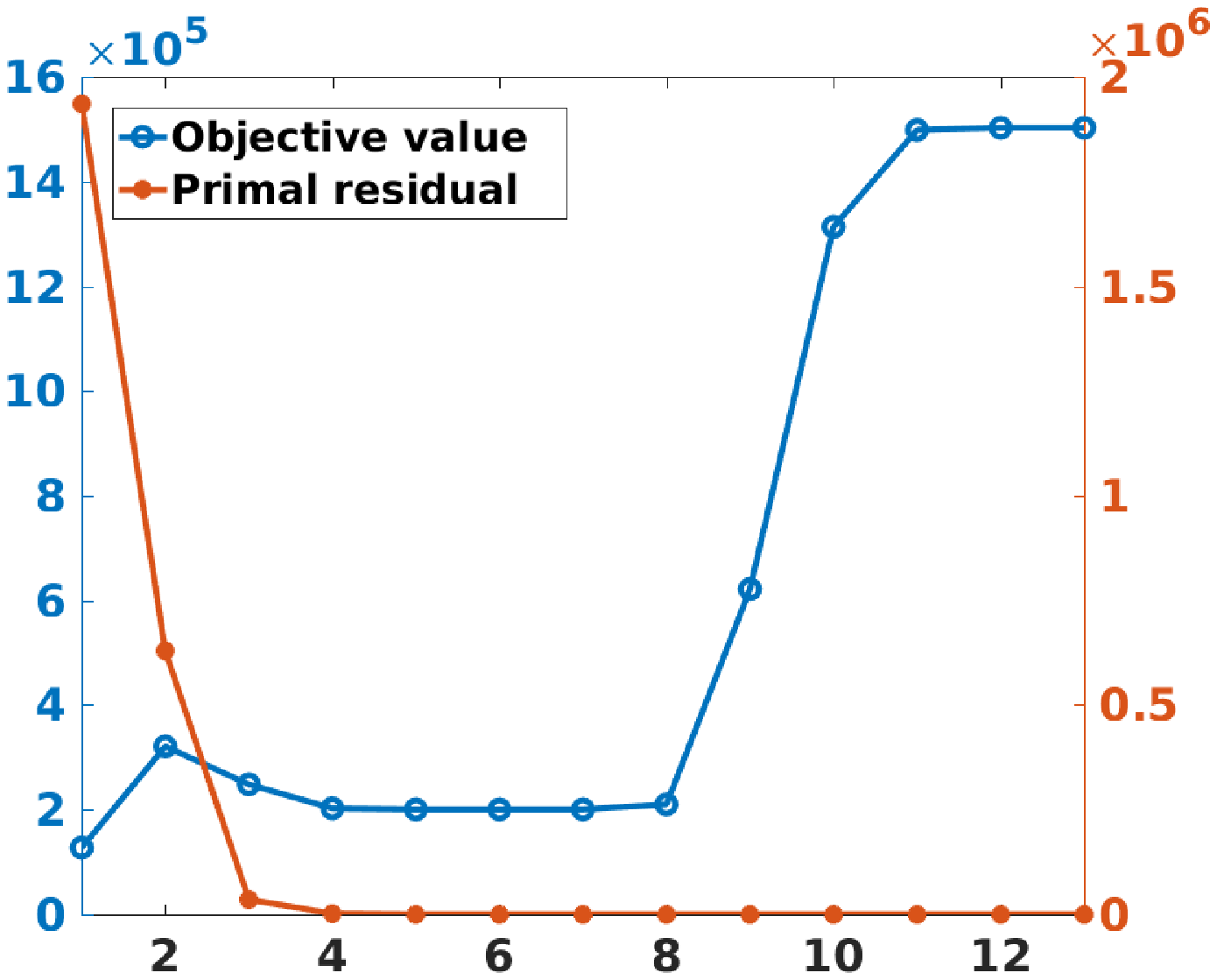}
\end{tabular}
\caption{Convergence curves for the objective values and primal residuals. {\bf Left:} convergences curves for Algorithm~\ref{Algorithm1}; {\bf right:} convergences curves for Algorithm~\ref{Algorithm2}.}
\label{fig:convergence}
\end{figure}

In this part of the experiments, we examine how Algorithm~\ref{Algorithm1} and Algorithm~\ref{Algorithm2} converge. Specifically, we compute the objective values of \eqref{eq:lrksc3} and  \eqref{eq:lrksc4}, and their primal residuals at each iteration. The primal residuals are respectively computed as $\min(\|{\bf A} - {\bf C} + {\rm diag}({\bf C})\|_F, \|{\bf 1}^T{\bf C} - {\bf 1}^T\|_2)$ for Algorithm~\ref{Algorithm1}, and $\min(\|{\bf A} - {\bf C} + {\rm diag}({\bf C})\|_F, \|{\bf 1}^T{\bf C} - {\bf 1}^T\|_2, \|{\bf K}_G - {\bf B}^T{\bf B} - {\bf E}\|_F)$ for Algorithm~\ref{Algorithm2}. We show typical convergence curves in Figure~\ref{fig:convergence} with data sampled from Hopkins155 and Extended Yale B. We can see that both algorithms converge fast (within 15 iterations) with the primal residuals quickly reduced close to zero. This is mainly due to our use of a relatively large $\eta$ in the ADMM, which gives rise to a large penalty parameter $\rho$ in a few iterations 
and thus greatly accelerates convergence. As we will see in the following sections, the solutions obtained by the ADMM are good in the sense that the corresponding results outperform the state-of-the-art on multiple datasets.

\subsection{Motion Segmentation on Hopkins155}

Hopkins155~\cite{tron2007benchmark} is a standard motion segmentation dataset consisting of 155 sequences with two or three motions. The sequences can be divided into three categories, \ie, indoor checkerboard sequences (104 sequences), outdoor traffic sequences (38 sequences), and articulated/non-rigid sequences (13 sequences). This dataset provides ground-truth motion labels, and outlier-free feature trajectories (x-, y-coordinates) across the frames with moderate noise. The number of feature trajectories per sequence ranges from 39 to 556, and the number of frames from 15 to 100. Since, under the affine camera model, the trajectories of one motion lie on an affine subspace of dimension up to three, subspace clustering methods can be applied for motion segmentation.

\begin{table}[!t] 
\centering
\small
\caption{\normalsize Clustering error (in \%) on Hopkins155.}
\label{tab:hp155}
\hspace*{-0.1cm}\begin{tabular}{ | l | c | c | c | c | c | c |}
\hline
  Method      & LRR  &  SSC & KSSC & LRSC  & {\footnotesize SSC-OMP} & Ours \\
            \hline\hline
  \multicolumn{7}{l}{\bf 2 motions}\\
  \hline
  Mean        & 2.13 & 1.53 & 1.85 & 2.57  & 10.34   & {\bf 1.22} \\
  Median      & {\bf 0.00} & {\bf 0.00} & {\bf 0.00} & {\bf 0.00}  & 3.48    & {\bf 0.00}\\
  \hline
  \multicolumn{7}{l}{\bf 3 motions} \\
  \hline
  Mean        & 4.13 & 4.41 & 3.57 & 6.64  & 18.58   & {\bf 3.10} \\
  Median      & 1.43 & 0.56 & {\bf 0.30} & 1.76 & 13.01 & 0.56\\
  \hline
  \multicolumn{7}{l}{\bf All} \\
  \hline
  Mean        & 2.56 & 2.18  & 2.24 & 3.47  & 12.20   & {\bf 1.64}\\
  Median      & {\bf 0.00} & {\bf 0.00} & {\bf 0.00} & 0.09 & 5.11 & {\bf 0.00}\\
 \hline
\end{tabular}
\end{table}

The parameters for our method (with Formulation~\eqref{eq:lrksc3} solved using Algorithm~\ref{Algorithm1}) are set to $\lambda_1 = 1$, $\lambda_2 = 12.6$, $\lambda_3 = 1\times 10^5$, and we use the polynomial kernel $\kappa({\bf x}_1, {\bf x}_2) = ({\bf x}_1^T{\bf x}_2 + 2.2)^3$ to define ${\bf K}_G$ in our formulation. We use the same polynomial kernel for KSSC. Since the input subspaces are affine but the affine constraint in~\eqref{eq:lrksc3} is in feature space, we append an all-one row to the data matrix, which acts as an affine constraint for the input space. This trick is also applied for KSSC. For the other baselines, we either use the parameters suggested in the original papers (if the parameters were given therein), or tune them to the best. We show the results on the Hopkins155 motion segmentation dataset in Table~\ref{tab:hp155}. Since most sequences in this dataset fit well to an affine camera model, most of the baselines perform well. Note that our method still achieves the lowest clustering errors, whereas KSSC performs slightly worse than SSC. The performance gain of our method over SSC mainly comes from the ability to handle the nonlinear structure that occurs when the affine camera model assumption is not strictly fulfilled. For example, in Figure~\ref{fig1:2RT3RCR}, we show the results on the 2RT3RCR sequence, which contains noticeable perspective distortion, and our method performs significantly better than SSC.

We further test our method for two-frame perspective motion segmentation~\cite{li2013perspective,ji2016robust} on Hopkins155 to rule out the effects of inaccurate camera model assumptions. For two perspective images, the subspace comes from rewriting the epipolar constraint~\cite{hartley2003multiple} ${{\bf x}'}^T{\bf F}{\bf x} = 0$, where ${\bf F}$ is the fundamental matrix, and ${\bf x}' = [x',y',1]^T$, ${\bf x} = [x, y, 1]^T$ are the homogeneous coordinates of two correspondence points. The epipolar constraint can be rewritten as~\cite{ji2016robust}
\begin{equation}
{\bf f}^T{\rm vec}({\bf x}'{\bf x}^T) = 0\;,
\end{equation}
where ${\bf f}$ is the vectorization of the fundamental matrix ${\bf F}$. So ${\rm vec}({\bf x}'{\bf x}^T)$ lies on the epipolar subspace (\ie, orthogonal complement of ${\bf f}$) of dimension up to eight~\cite{ji2016robust}. Since different motions correspond to different fundamental matrices, we have multiple epipolar subspaces for two perspective images with multiple motions.

We take the first and last frames from each sequence of Hopkins155 to construct the two-frame Hopkins155 dataset. Note that the dimension of the ambient space of epipolar subspaces is only nine, so the epipolar subspaces are very likely to be dependent. To increase the ambient dimension, we replicate the data 30 times\footnote{In Matlab, this is done by repmat(${\rm vec}({\bf x}'{\bf x}^T)$,30,1).}. We set the parameters of our method (with Formulation~\eqref{eq:lrksc3} solved using Algorithm~\ref{Algorithm1}) on two-frame Hopkins155 as $\lambda_1 = 0.23$, $\lambda_2 = 5.5$, $\lambda_3 = 1\times 10^5$, with ${\bf K}_G$ the polynomial kernel $\kappa({\bf x}_1, {\bf x}_2) = ({\bf x}_1^T{\bf x}_2 + 2)^2$. The results are shown in Table~\ref{tab:hp155twoframe}, where our method achieves the lowest overall clustering errors. Note that our method with only two frames even outperforms LRSC with the whole sequences.

\begin{table}[!t]
\centering
\small
\caption{\normalsize Clustering error (in \%) on two-frame Hopkins155.}
\label{tab:hp155twoframe}
\hspace*{-0.25cm}\begin{tabular}{ | l | c | c | c | c | c | c |}
\hline
  Method      & LRR  &  SSC & KSSC & LRSC  & {\footnotesize SSC-OMP} & Ours \\
            \hline\hline
  \multicolumn{7}{l}{\bf 2 motions}\\
  \hline
  Mean        & 6.18 & 3.52 & 2.75 & 3.74  & 14.41   & {\bf 2.53} \\
  Median      & 0.10 & 0.10 & {\bf 0.00}  & 0.85  & 9.60   & 0.36\\
  \hline
  \multicolumn{7}{l}{\bf 3 motions} \\
  \hline
  Mean        & 14.92 & 10.93 & 8.60  & 22.28 & 37.32   & {\bf 5.93} \\
  Median      & 10.16 & 8.65  & 5.40  & 20.72 & 39.17   & {\bf 0.20}\\
  \hline
  \multicolumn{7}{l}{\bf All} \\
  \hline
  Mean        & 8.15 & 5.20 & 4.07 & 7.93  & 19.58   & {\bf 3.30}\\
  Median      & 0.54 & 0.52 & 0.40 & 1.81  & 15.98 & {\bf 0.36}\\
 \hline
\end{tabular}
\end{table}

\subsection{Face Image Clustering on Extended Yale B}

\begin{figure*}[!t]
\includegraphics[width=1.00\linewidth]{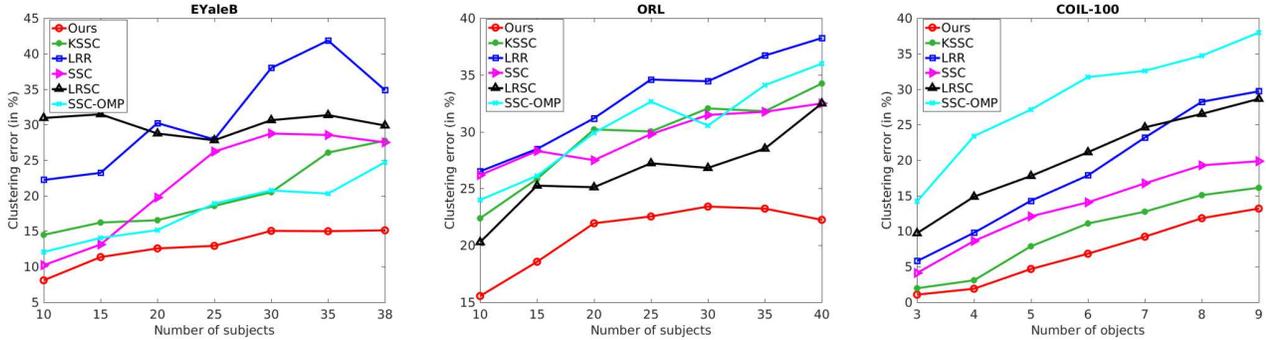}
\caption{Mean clustering errors (in \%) on Extended Yale B (EYaleB), ORL, and COIL-100 w.r.t. different numbers of clusters.}
\label{fig:result_all}
\end{figure*}

The Extended Yale B Dataset~\cite{lee2005acquiring} consists of aligned face images of 38 subjects. Each subject has 64 frontal face images, which are acquired under a fixed pose with varying lighting conditions. It has been shown in~\cite{basri2003lambertian} that, under Lambertian reflection, the images of a subject under the same pose with different illuminations lie close to a 9-dimensional linear subspace. Following~\cite{elhamifar2013sparse}, we down-sample the face images to size $48\times 42$, and vectorize them to form 2016-dimensional vectors. Each 2016-dimensional image vector lies close to a low-dimensional subspace. The dataset contains sparse gross corruptions due to the existence of specular reflections, which are non-Lambertian. Non-linearity arises because the face poses and expressions were not exactly the same when images were taken for the same subject.

We test our method and the baselines on this dataset with different numbers of subjects ($K=$ 10, 15, 20, 25, 30, 35, or 38). We number the subjects from 1 to 38. We first take the first $K$ subjects, and then take the next $K$ subjects until we consider all of them. For example, for 10 subjects, we take all the images from subjects 1-10, 2-11, $\cdots$, or 29-38 to form the data matrix ${\bf X}\in\mathbb{R}^{2016\times 640}$ for each trial.

We use Formulation~\eqref{eq:lrksc4} (solved using Algorithm~\ref{Algorithm2}) for our method, and set the parameters to $\lambda_1 = 1.1\times10^3$, $\lambda_2 = 2\times 10^{-2}$, $\lambda_3 = 1\times 10^5$, with ${\bf K}_G$ the polynomial kernel $\kappa({\bf x}_1, {\bf x}_2) = ({\bf x}_1^T{\bf x}_2 + 12)^2$. For our method, we normalize the data to lie within $[-1,1]$. We use the same polynomial kernel for KSSC. The results on Extended Yale B are shown in Figure~\ref{fig:result_all}. We can see from the table that KSSC improves the clustering accuracies over SSC for 20, 25, 30, and 35 subjects. Our low-rank kernel subspace clustering method achieves the lowest clustering errors on this dataset for all numbers of subjects. For 20, 25, 30, 35 and 38 subjects, our method almost halves the clustering errors of SSC, and also significantly outperforms all baselines including KSSC.

\subsection{Face Image Clustering on the ORL Dataset}

The ORL dataset~\cite{samaria1994parameterisation} is composed of face images of 40 distinct subjects. Each subject has ten different images taken under varying lighting conditions, with different facial expressions (open/closed eyes, smiling/not smiling) and facial details (glasses/no glasses). Following~\cite{cai2007learning}, we crop the images to size $32\times 32$, and then vectorize them to 1024-dimensional vectors. We use a similar experimental setting as for Extended Yale B, and test the algorithms with different numbers of subjects ($K=$ 10, 15, 20, 25, 30, 35, 40). Compared to Extended Yale B, ORL has fewer images for each subject (10 vs. 64), higher subspace non-linearity due to variations of facial expressions and details, and is thus more challenging.

We use Formulation~\eqref{eq:lrksc4} (solved using Algorithm~\ref{Algorithm2}) for our method, and set the parameters to $\lambda_1 = 1\times10^3$, $\lambda_2 = 6\times 10^{-2}$, $\lambda_3 = 1\times 10^5$, with ${\bf K}_G$ the polynomial kernel $\kappa({\bf x}_1, {\bf x}_2) = ({\bf x}_1^T{\bf x}_2 + 12)^2$. For our method, we normalize the data to lie within $[-1,1]$. We use the same polynomial kernel for KSSC. We show the results on the ORL dataset for all competing methods in Figure~\ref{fig:result_all}. We can see that KSSC cannot outperform SSC for 20, 25, 30, 35 and 40 subjects on this dataset. We conjecture that this is mainly because for each subject we only have ten images, which are too few to span the whole space. Our superior performance verifies that, by adaptively solving for a low-rank feature mapping, our method can better handle this ``very-few-sample'' case. 

\subsection{Object Image Clustering on COIL-100}


The COIL-100 dataset~\cite{nenecolumbia} contains images of 100 objects. Each object has 72 images viewed from varying angles. Following~\cite{cai2011graph}, we down-sample them to $32\times 32$ grayscale images. Each image is vectorized into a 1024-dimensional vector, which corresponds to a point lying in a low-dimensional subspace. As in the previous experiment on Extended Yale B, we also test our method and the baselines on this dataset for different numbers of objects with $K = $ 3, 4, 5, 6, 7, 8, or 9. For $K = 3$ (\ie, three objects), we take all the images from objects 1-3, 2-4, $\cdots$, 98-100 to form the data matrix ${\bf X}\in\mathbb{R}^{1024\times 216}$ for each trial. The data matrices for the other $K$s are formed in a similar manner.

Again, we use Formulation~\eqref{eq:lrksc4} (solved using Algorithm~\ref{Algorithm2}) for our method, and set the parameters to $\lambda_1 = 1.4\times10^3$, $\lambda_2 = 6\times 10^{-2}$, $\lambda_3 = 1\times 10^5$, with ${\bf K}_G$ the polynomial kernel $\kappa({\bf x}_1, {\bf x}_2) = ({\bf x}_1^T{\bf x}_2 + 12)^2$. We also use the same kernel for KSSC. We show the results on COIL-100 in Figure~\ref{fig:result_all}. We can see that KSSC consistently outperforms SSC in this setting, which indicates that there is a considerable amount of non-linearity in this dataset. Our method, by solving a low-rank kernel mapping, achieves the lowest clustering errors among all the baselines.

\paragraph{Discussion:} For all our experiments, we do not mean to claim that our results are the best among all the literature, but to showcase that our adaptive low-rank kernel subspace clustering improves over its linear counterpart and the kernel method that uses fixed kernels. For example, we note that, on Hopkins155, better results were reported by~\cite{jung2014rigid,ji2015shape}. However, to the best of our knowledge, our method is the first kernel subspace clustering method that achieves better results than its linear counterpart on Hopkins155 where most of the data conforms to the linear subspace structure very well.

\section{Conclusion}

In this paper, we have proposed a novel formulation for kernel subspace clustering that can jointly optimize an adaptive low-rank kernel and pairwise affinities between data points (through subspace self-expressiveness). Our key insight is that instead of using fixed kernels, we need to derive a low-rank feature mapping such that we have the desired linear subspace structure in the feature space. We have derived efficient ADMM solutions to the resulting formulations, with closed-form solutions for each sub-problem. We have shown by extensive experiments that the proposed method significantly outperforms kernel subspace clustering with pre-defined kernels and the state-of-the-art linear subspace clustering methods.

The main limitation of the current method is that we still need to manually select a kernel function to construct ${\bf K}_G$. 
In the future, we plan to explore the possibility of employing the multiple kernel learning method~\cite{gonen2011multiple} to automatically determine ${\bf K}_G$.

{\small
\bibliographystyle{ieee}
\bibliography{subspaceclustering}
}

\end{document}